\documentclass[fleqn,11pt]{wlscirep}
\usepackage[parfill]{parskip}
\usepackage{float}
\usepackage{graphicx}
\PassOptionsToPackage{hyphens}{url}\usepackage{hyperref}

\graphicspath{{figures/}}
\title{A Cyber Science Based Ontology for Artificial General Intelligence Containment}

\author[1,*]{Jason M. Pittman}
\author[1]{Courtney E. Soboleski}

\affil[1]{Capitol Technology University, Synthetic Intelligence Research Institute , Laurel, MD, USA}
\affil[*]{corresponding author}

\begin{abstract}
The development of artificial general intelligence is considered by many to be inevitable. What such intelligence does after becoming \textit{aware} is not so certain. To that end, research suggests that the likelihood of artificial general intelligence becoming hostile to humans is significant enough to warrant inquiry into methods to limit such potential. Thus, containment of artificial general intelligence is a timely and meaningful research topic. While there is limited research exploring possible containment strategies, such work is bounded by the underlying field the strategies draw upon. Accordingly, we set out to construct an ontology to describe necessary elements in any future containment technology. Using existing academic literature, we developed a single domain ontology containing five levels, 32 codes, and 32 associated descriptors. Further, we constructed ontology diagrams to demonstrate intended relationships. We then identified \textit{humans}, \textit{AGI}, and the \textit{cyber world} as novel agent objects necessary for future containment activities. Collectively, the work addresses three critical gaps: (a) identifying and arranging fundamental constructs; (b) situating AGI containment within cyber science; and (c) developing scientific rigor within the field. 
\end{abstract}
\begin{document}

\flushbottom
\raggedbottom
\maketitle

\thispagestyle{empty}

\section*{1. Introduction}

Artificial intelligence (AI) is a timely and germane topic. No longer the esoteric domain of academics and researchers alone, the public is well aware of the extensive progress in the field. Mainstream media has brought practical applications such as autonomous vehicles (e.g. Tesla) and leisure game victories (e.g., Waston, AlphaGo) to the forefront of daily life. Thinking machines are a realization of science fiction, writ large. Certainly, the positive aspects of continuing to develop and deploy such systems are disruptive (Sotala \& Yampolskiy, 2016). However, such \textit{narrow AI} operates in limited operational scope and presents little direct risk to human life. On the other hand, a synthetic intelligence with the ability to dynamically operate in general scopes may indeed present direct and sustained danger.

Artificial general intelligence (AGI) is precisely that; AI with intelligence at or beyond human capability. While the concept of AGI is not new, there has been rapid growth in the literature surrounding the subject. AGI, perhaps unfortunately, is framed broadly in two manners: (a) in popular culture, as something that is simultaneously inevitable (Amodei et al., 2016; Baum, Goertzel, \& Goertzel, 2011; Kurzweil 2005) and (b) of existential danger (Bostrom, 2002; Bostrom \& Yudkowsky, 2011; Bostrom \& Cirkovic, 2011). The nexus of these two frames has engendered serious research into ethics and transhumanism, as well as safety and trust.  

This study focuses on a specific area of research in the latter category: \textit{containment} of AGI. According to Babcock, Kramar, \& Yampolskiy (2016), viable AGI containment will necessitate a combination of traditional cybersecurity technologies such as (a) safe language semantics; (b) physical separation of systems; (c) sandboxing; and (d) virtualization. However, Babcock et al. recognized the problem of limited capacity for traditional cybersecurity paradigms to address AGI containment. That is, the technologies Babcock et al. suggested may present a small window into what the broader field of cyber science can offer to future containment research and practical development. 

Cyber science is a nascent field that considers cybersecurity in a more comprehensive knowledge context (Ma, Nahal, \& Tran, 2015; Maxion, Longstaff, \& McHugh, 2010; McDaniel, Rivera, \& Swami, 2014). An applicable sector within the cyber science was defined by Kott (2015) as principally considering malicious software. The applicability rests in the nuance of Kott's argument; that is, "malicious software (as well as legitimate software and protocols used maliciously) used to compel a computing device or a network of computing devices to perform actions desired by the perpetrator of malicious software (the attacker) and generally contrary to the intent (the policy) of the legitimate owner or operator (the defender) of the computing device(s)" (pg. 1). While there could be debate as to whether an AGI is malicious software (in the perspective of humans) or a containment apparatus is malicious software (in the perspective of the AGI), there is not a means to organize the underlying constructs that engender meaning.   

Thus, we endeavored to build upon the initial discussion initiated by Babcock et al. (2016) by developing a cyber science focused ontology for AGI containment. The significance in doing so rests in the necessity of having a formalized, systematic abstraction from which future research can construct scientific inquiry. Moreover, such an ontology may fill existing gaps in a manner that enables future applied research in AGI containment.

\section*{2. Background}

Scientists have sought to engineer AI with intelligence at or beyond human capability by aggregating existing narrow AI technology. Narrow AI is an implementation tool for enhancing human tasks in limited domains (Stone et al., 2016). The problem is that narrow AI cannot be aggregated to birth general AI (Baillie, 2016). Accordingly, the purview has transitioned from manufacturing singular intelligent parts to conceiving entire intuitive systems. Such systems have instigated innovations in novel training approaches (Guo \& Aarabi, 2016), learned systems for internal efficiency (Le \& Schuster, 2016), greater contextual and environmental awareness (Denil et al., 2017), and emergence (Silver et al., 2016). Progress has been further accelerated by advancements in big data, machine learning, and computer processing (Stone et al., 2016).

The result is not only a catalogue of technologies capable of functioning appropriately with environmental foresight (Nilsson, 2009) but an entirely new cyber science. While, artificial general intelligence portends the ability of an intelligent machine to infer, reason, and adapt to its environment, cyber science contextualizes such AGI within a more comprehensive knowledge framework (Kott, 2015). Cyber science contrasts existing literature that rationalizes AGI solely within the physical world of humans (Bostrom, 2002; Bostrom \& Yudkowsky, 2011; Amodei et al., 2016). Instead, cyber science trains the lens of AGI investigation on the complex and unpredictable cyber world a future entity may exist in (Oltramari, Cranor, Walls, \& McDaniel, 2014; Ma et al., 2015; McDaniel et al., 2014). Accordingly, actuating AGI is constrained by the implications of unpredictability (Agar, 2016; Stone et al. 2016). In fact, an entire subfield is dedicated to mitigating potential AGI through containment (Babcock, Kramar, \& Yampolskiy, 2017).

Containment seeks to avert scenarios where AI maliciously competes against humans through strict regulation and control (Bostrom, 2014; Babcock et al., 2016). Conditions include not only technical protocols but ethical, moral, and legal considerations as well (Yampolskiy, 2012; Powers, 2006; Lewis \& Modirzadeh, 2016). Such controls are predicated on the assumption that AGI not only possesses a catastrophic, even existential, risk to human society but that it will act on such risk by intentionally harming humans (Müller, 2014; Bostrom, 2014; Sotala \& Yampolskiy, 2014; Chalmers, Awret, \& Appleyard, 2016; Amodei et al., 2016; Yudkowsky, 2008). Accordingly, containment literature demonstrates a singular preference toward delaying or banning AI advancements altogether to achieve a zerosum, human favored reality (Babcock, et al., 2017). Yet, existing containment literature is plagued by myriad definitions and paradigms, an assimilation of danger through an anthropomorphic lens, and a focus on avoidance that assumes a negative position of inevitability (Bishop, 2009; Grau, 2006; Yampolskiy \& Fox, 2012). Thus, implementation of containment policies is not only inherently obtuse but contradictory to the entire premise of superintelligence.

Agar (2016) challenged containment assumptions by arguing that solutions to the AGI problem are as conceivable as AGI itself. In fact, human society will become increasingly more adept at handling AGI as continued artificially intelligent innovations become common place (Agar, 2016; Stone et al. 2016). That is, Agar and Stone assert that if humans can develop a superintelligent machine, then we are equally as capable of harnessing such technology to solve any subsequent issues that arise from it. Developing an architecture that accurately accounts for such an assertion requires containment be mapped to the greater cyber theoretical framework (Ning, et al., 2017). Such an architecture is the crux of cyber science, which supplements technical and physical premises with philosophical and social ones (Maxion et al., 2010; Oltramari et al., 2014; Ning, 2016).

Yet, while an ontological architecture for AGI containment been suggested (Yampolskiy \& Fox, 2012; Yampolskiy, 2014, 2016), one has never been fully codified (Amodei et al., 2016). As a result, the cyber community lacks a measurable or repeatable standard from which to understand or predict cyber events and entities (Maxion et al., 2010). An ontology for AGI containment therefore, accounts for the broader cyber context by systematically converting existing AGI abstractions into defined relationships (Kott, 2015). In doing so, our cyber sciencefocused ontology for AGI observes domain constructs, remains iterative, and applies a level of scientific rigor currently underdeveloped within the discipline (Uschold \& King, 1995; Ren, 2012; Maxion et al., 2010). Simultaneously, our analysis will leverage the same multidisciplinary approaches compulsory in AGI development to its germinal research. The resulting ontology may establish AGI as a cyber science domain and enable future empirical research.

\section*{3. Method}

A single research question motivated this study: what specific concepts and relationships between such concepts are necessary so that AGI containment functionality can be explored? Toward an answer to this research question, we endeavored to develop a cyber science focused ontology for AGI containment. Development of an ontology appeared appropriate because, as Babcock et al. (2016, 2017) commented, there is uncertainty as to how AGI containment ought to function. Further, as a scientific field defined by both theoretical and empirical methods, exploring cyber science relationships requires a foundational ontology (Longstaff, 2010). In other words, an ontology can mitigate conceptual misunderstanding.

Poli (1996) suggested that ontologies provide generalized frameworks in which more specific knowledge taxonomies can be developed. Further, as Poli noted, the unavailability of an ontology limits the relational context for future inferences or knowledge construction. Indeed, the lack of an ontology inhibits meaningful discourse not only concerning concepts but also with applied constructs. Accordingly, we suspected that a lack of foundational ontology might be a cause for the uncertainty related to AGI containment described by Backcock et al (2016, 2017). Thus, to formally represent the concepts that ought to be included in AGI containment, we pursued a single domain ontology (Subhashini \& Akilandeswari, 2011).

\subsection*{3.1 Method Appropriateness}

We considered a grounded theory design for our research methodology. Grounded theory is appropriate when one seeks to develop a theory based on emergent themes in data (Martin \& Turner, 1986). That is, grounded theory produces an explanation of \textit{what is} given existing qualitative data. Such an explanation is achieved by iteratively collecting and analyzing data. However, according to Suddaby (2006), the results from grounded theory research are limited by what is known. Thus, while there may be some notional overlap between grounded theory and ontology, the former presupposes the latter. As well, we did not set out to facilitate the output of theory as much as the underlying conceptual vocabulary to support future theoretical frameworks.

Existing research (Subhashini \& Akilandeswari, 2011; Swartout, Patil, Knight, \& Russ, 1997) outlined two toplevel forms of ontology: domain and theory. Domain ontologies are classlevel vocabularies that express concepts and relationships between those concepts (Swartout et al., 1997). In contrast, Swartout et al. defined theory ontologies as describing aspects of (our) reality. While a theory ontology could be used, we decided that the reality of an AGI in a containment space would be best described by that AGI. Whereas, we felt developing concepts related to the containment space itself to be a rational position.

Additionally, we elected to develop a single domain ontology as opposed to a multiple domain ontology or a hybrid ontology. Relevant literature (Subhashini \& Akilandeswari, 2011; Swartout, Patil, Knight, \& Russ, 1997) indicated that a single ontology best establishes a global ontology based on the collection of concept sources. The single ontology approach appeared most fitting relative to our purpose of developing a universal ontology for AGI containment. In contrast, a multiple ontology or hybrid ontology approach did not align with our objective due to those methodologies necessitating the development of local ontologies mapped individually to the source material.

\subsection*{3.2 Existing Ontologies}

Before developing our ontology, we performed a literature review to ensure that no ontology existed which addressed our research question. This literature search was distinct from that performed later as a means of collecting literature for the development of the AGI containment ontology proposed in this work. While a rich knowledge base exists for narrow aspects of artificial intelligence (e.g., natural language processing), we did not find any ontologyrelated AGI containment literature. The lack of such research was not surprising given the nascent form of the AGI containment field and does not reflect negatively on the existing work on the topic. Next, we searched for germane literature associated with cyber science and cybersecurity.

Interestingly, both cyber science and cybersecurity offered foundational ontologies albeit without any direct relevance to AI, AGI, or AGI containment. Moreover, according to Oltramari, Cranor, Walls, and McDaniel (2014), existing cybersecurity ontologies (Blanco et al., 2008) are flawed. Further, attempts to achieve a robust cyber science ontology have fallen short (Maxion et al., 2010; Ning, 2017). Although this study does not address the limitations of cyber science or cybersecurity ontologies, we did ensure that such were not carried forward into our proposed ontology.

\subsection*{3.3 Ontology Construction Methodology}

Two studies directly informed how we developed our ontology. Foremost, we consulted the seminal work by Uschold and King (1995) for a general sketch of what phases we should follow when constructing a new ontology. Further, to minimize potential errors while determining fundamental AGI containment concepts, we leveraged the ontology building process outlined by Ren (2012). Ren suggested that the use of academic literature as the foundational source of concepts presented a more reliable and efficient ontology building process. Not only did literature as a material concept source present a sound rationale for this work but such also conformed to our stated purpose through enabling the single domain methodology.

Furthermore, our ontology construction process focused on the designing and development steps as described by Subhashini and Akilandeswari (2011). Integrating the new ontology with existing ontologies was not necessary for obvious reasons. Additionally, given the germinal nature of this work, we opted to postpone refinement of the proposed ontology through validation and iteration. 
Subordinate to the design and develop steps, we employed the specific construction phases outlined by Ren (2012). That is, we (a) collected academic literature associated with AGI containment, (b) selected potential ontology concepts based on keywords in the literature, and (c) extracted relationships between such potential concepts based on keyword frequencies. The relationships between concepts were mapped using two graph morphologies.

\section*{4. Results}

We opted for a modular, tiered ontology architecture comprised of objects, attributes, and relationships (Oltramari et al., 2014; Subhashini \& Akilandeswari, 2011; Ren, 2012). Concurrently, an axiomatic approach created a foundation from which our ontology could extract further reason around said objects, attributes, and relationships. Axioms are stated along with the ontology element they describe. The structure of our ontology is detailed as follows.

\subsection*{4.1 Objects}

Objects (Table 1) in our cyber science ontology for AGI containment represent upper domain areas. As an upper domain area, objects are agents through which AGI containment occurs. Thus, without such agents, containment could not occur. Furthermore, agents can be understood as entities, of which there are three types: humans, AGI, and the cyber world. The three agents are discrete, meaning that a human \textit{cannot be} AGI \textit{cannot be} a cyber world. 

\begin{table}[H]
\centering
\begin{tabular}{|r|ccc|}
\hline
Object Coding & O1 & O2 & O3 \\ 
Object Descriptor & human & AGI & cyberworld \\
\hline
\end{tabular}
\caption{Tiered ontology description of levels, codes, and descriptors for Objects}
\label{table:ontological-objects}
\end{table}
\vspace*{-10pt}
Humans are an organic agent capable of autonomy and intelligence whereas AGI are an artificial agent capable of autonomy and intelligence. Indeed, the capacity for autonomy and intelligence is fundamental to containment because without such features, agents would have no need, desire, or intent. Thus, the final agent of containment is the cyber world, which is the digitized environment in which organic human and artificial general intelligence agents exist (Ma et al., 2015). That is, containment can only exist in the cyber world, and never occurs in a conventional, noncyber enabled world.

\begin{table}[H]
\centering
\begin{tabular}{|r|ccccccc|}
\hline
Class Coding & C1 & C2 & C3 & C4 & C5 & C6 & C7 \\ 
Class Descriptor & individual & society & swarm & physical & social & mental & cyber \\
\hline
\end{tabular}
\caption{Tiered ontology description of levels, codes, and descriptors for Classes}
\label{table:ontological-classes}
\end{table}
\vspace*{-10pt}

Agents are concentrated based on their instance and class (Table 2): a single human or AGI instance is an individual, while a class of humans is a society and a class of AGI is a swarm. Similarly, the cyber world is composed of four class concepts: the cyber, physical, social, and mental (CPSM) (Ning, 2017). Instance and class serve to further subcategorize and delimit agent relationships. For example, a distinction is drawn between each cyber world class since humans and AGI are not required to exist in all four simultaneously. Each instance and class also exhibit and maintains their own features. Specifically, the cyber realm is "anything that exists digitally in cyberspace, either purely synthesized by a computer, or closely correlated to and further conjugated with a real entity in physical, social and mental spaces" (Ma et al., 2015). Meanwhile, the physical, social, and mental space mimic those of conventional worlds but are understood from a cyber science perspective as conjugations of objects. In this case, subcategorization arranges the ontology hierarchically (Figure 1) to more easily segregate objects.

\begin{figure}[H]
\centering
\includegraphics[trim=4 4 4 4,clip]{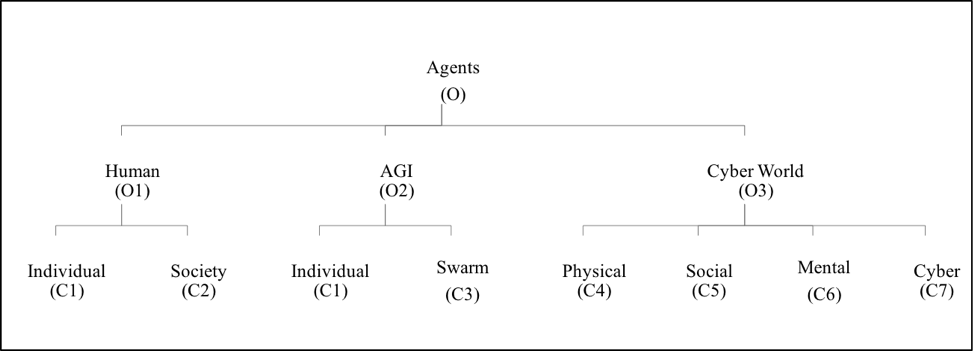}
\caption{Diagram of agent objects (O), their classes(C) in the AGI containment ontology.}
\label{fig:1}
\end{figure}

\subsection*{4.2 Attributes}

Another form of segregation occurs through physical and abstract attributes (Table 3). Physical attributes provide the infrastructure from which agents compose and leverage capabilities. Thus, physical attributes enable delineation by creating a unique signature summarized by the composition, architecture, and locality of the agent.

\begin{table}[h]
\centering
\begin{tabular}{|r|cc|}
\hline
Atrribute Coding & A1 & A2 \\ 
Atrribute Descriptor & physical & abstract \\
\hline
\end{tabular}
\caption{Tiered ontology description of levels, codes, and descriptors for Attributes}
\label{table:ontological-attributes}
\end{table}

The object's signature separates one agent from another. For example, the nature of the agent (composition) and the structure of the agent (architecture) couple with locality to output a unique persona that marks an agent in space and time. Subcategorization in this sense serves to further distinguish an entity by creating a discrete identity. For instance, composition is the construction of matter (organic or inorganic) and visibility (able to be seen or not). Similarly, physical architecture categorizes the agent based on their hardware (the physical tools, implements, and parts used by entities) and software (the programs, libraries, and data used by entities for executing relationships). With locality providing a temporal and spatial reference point, physical attributes enable cyber science agent differentiation through unique traits (Figure 2). 

In addition to physical attributes, objects within the cyber science interpretation of AGI containment can be containerized by abstract attributes (Figure 3). In fact, while physical attributes ascribe identity, abstract attributes account for the concepts that predicate a containment environment. Thus, abstract attributes are the capabilities leveraged by objects in a cyber science world. Capabilities come in the form of security, intelligence, and autonomy. Regarding autonomy and intelligence, there are two axioms. There are no degrees of autonomy; an agent is autonomous or it is not. Additionally, intelligence is a physical process or the (emergent) consequence of one; not a metaphysical process or the (emergent) consequence of one (Wissner-Gross \& Freer, 2013).

\begin{figure}[H]
\centering
\includegraphics[trim=4 4 4 4,clip]{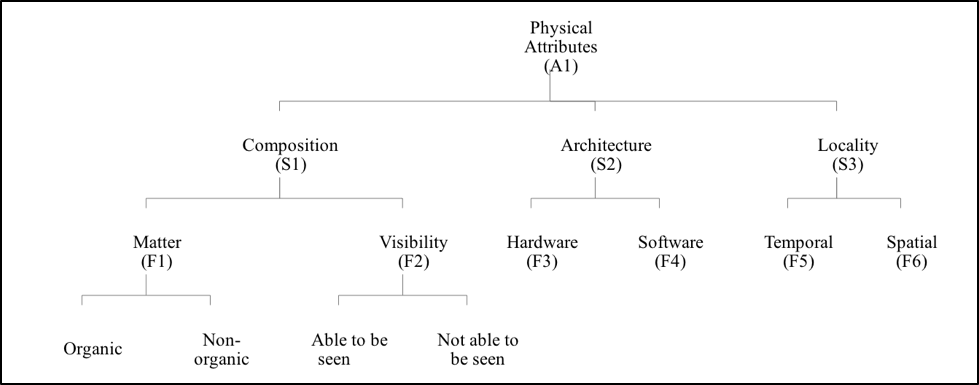}
\caption{Diagram of physical attributes (A1), subattributes (S), and their features (F) in the AGI containment ontology.}
\label{fig:2}
\end{figure}

Further, security is a phenomenon that exists within spectrums of logic and uncertainty, where logic is the essence of rules applied to cyber science objects and uncertainty is the unpredictability of cyber science objects (Kott, 2015). Conversely, intelligence and autonomy are dictated by binary properties, with intelligence being the capacity to acquire knowledge and skills and autonomy being the capacity to exercise independent control over one's own intelligence. Such binary properties serve to create discrete features. For example, there is no spectrum of quality or composition; the quality can only be narrow or general and the composition can only be artificial or organic (made by chemical synthesis) or artificial (not existing naturally). Such discrete features compliment degree-based properties to create a ho-listic abstraction of the agent.

\begin{figure}[H]
\centering
\includegraphics[trim=4 4 4 4,clip]{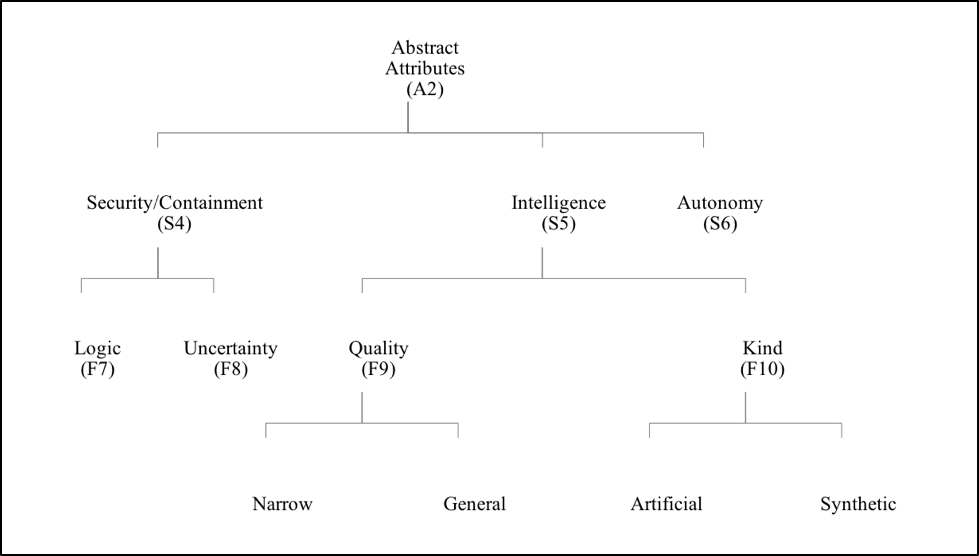}
\caption{Diagram of abstract attributes (A2), subattributes (S), and their features (F) in the AGI containment ontology.}
\label{fig:3}
\end{figure}

\subsection*{4.3 Relationships}

With defined agents and such agent identities, relationships serve to outline interaction between objects within the context of containment. The relationships described in our cyber science ontology (Figure 4) are either active or passive. Passive relationships enable the environment and consistent of existence (having objective reality) and policy (a set of assertions or requirements) (Kott, 2015). Without existence or policy, containment is an irrelevant non-factor. Yet, converting containment \textit{the concept} into containment \textit{the executable} requires that active relationships be applied to the cyber science world. Thus, the intent of active relationships is to either attack or defend. 

The dynamics of attack or defend relationships are such that all relationships occur within a cyber world; and active relationships occur only after passive ones. That is, agents must be in existence and policy must be violated for attack or defend activities to occur. Thus, relationships exhibit a sequential pattern: passive relationships occur; the cyber world is disturbed; the attacker employs tools or techniques to promote activity adverse to the intent of the defender; and the defender employs tools or techniques to re-initiate or establish their intent (Kott, 2015). Containment, therefore, can be described as an active relationship between agents, and within a cyber world, that (a) prevents an agent from disrupting the cyber world without authorization or (b) maintains the equilibrium achieved during active relationships (Babcock et al., 2016).  

\begin{figure}[H]
\centering
\includegraphics[trim=4 4 4 4,clip]{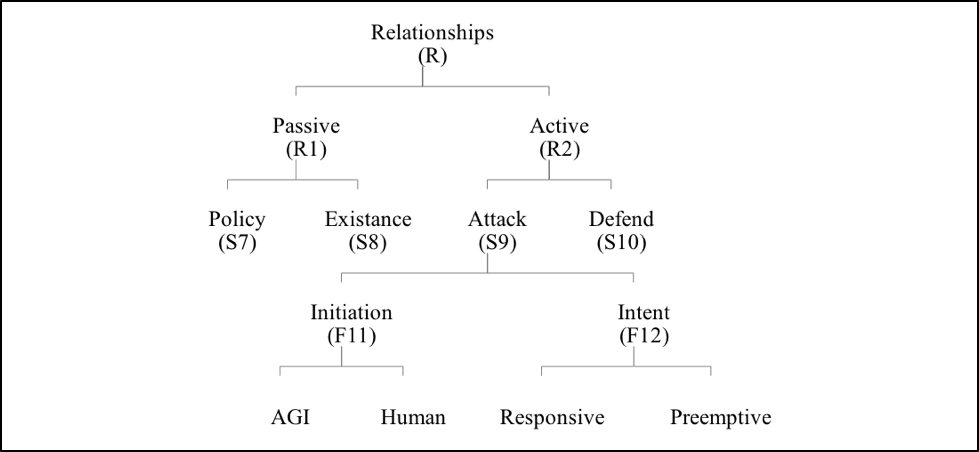}
\caption{Diagram of relationships (R), subattributes (S), and features (F) in the ontology for AGI containment.}
\label{fig:4}
\end{figure}

Given the function of containment \textit{f(c)} and current knowledge of the cyber world \textit{k}, both attacker and defender exist in a \textit{f(c) = k} environment. However, an attack implies that the attacker has acted on \textit{k}. Thus, the attack can be described as \textit{f(c) = k+1}, where \textit{+1} is the subsequent action of the attacker in response to \textit{k}. Thus, an attack has two features: initiation and intent. Initiation describes the agent acting on \textit{k} (human or AGI) while intent represents the catalyst for initiation \textit{(k)}. Intent therefore, is either preemptive or responsive, where preemptive is to prevent or forestall a perceived element of \textit{k} and responsive is a reaction to an actual element of \textit{k}. Attack features - initiation and intent - occur simultaneously. Conversely, a defense is always initiated by the non-attack agent. Initiation then, is singular rather than binary. Additionally, a defend relationship is always responsive since a preemptive defense is an attack.

\section*{5. Conclusions}

The cognitive leap from limited domain, narrow intelligence to super human AGI generated a novel discourse on the scope of future technologies (Horowitz, 2014; National Science and Technology Council, 2016). Indeed, the interest surrounding the prospect of super human intelligence led to a subsequent influx in literature that framed AGI largely in an imminent danger context (Müller, 2014; Bostrom, 2014; Sotala \& Yampolskiy, 2014; Chalmers et al., 2016; Amodei et al., 2016; Yudkowsky, 2008). The pretense of imminent danger led to the introduction of containment, or the application of traditional cybersecurity technologies to thwart AGI risks (Babcock et al., 2016). 

Containment research presupposes a binary world where AGI pits itself against all else to achieve a zerosum end state (Babcock et al., 2016; Bostrom, 2002; Bostrom \& Yudkowsky, 2011; Bostrom \& Cirkovic, 2011). As such, the literature fails to consider the full breadth and depth of the AGI context, while simultaneously ignoring common tenets from which to extract further meaning. Ironically, containment proponents themselves have suggested a standardized framework to enable comprehensive understanding and predict future events and entities (Yampolskiy \& Fox, 2012; Yampolskiy, 2014, 2016; Amodei et al., 2016). Yet, no such architecture had been brought forth. 

\section*{6. Recommendations}

The introduction of our cyber science AGI containment ontology addresses three critical gaps in the germinal research by (a) identifying and arranging underlying constructs; (b) placing containment squarely within the complex cyberworld of future AGI; and (c) establishing a level of scientific rigor previously underdeveloped within the discipline. Moreover, our architecture empowers other researchers with central axioms that enable replicability and shared understanding within the domain. 

Future AGI containment research may build upon the single domain ontology produced in this study. Specifically, because we have provided the means to organize the underlying constructs that engender meaning within AGI containment, our hope is that subsequent research begins to move away from the presupposed binary nature of AGI containment. Further, the ontology may aid future research investigating the malicious software perspective of AGI and containment.

Certainly, there is more work to be done. Whilst an ontology may provide a foundational basis for constructing new AGI containment knowledge, much work remains. Foremost, there is opportunity to produce theory ontologies that describe more detailed aspects of containment. Additionally, as the field of cyber science itself evolves, there is a need for a continuous integration between the converged knowledge domains involved in AGI containment. Lastly, there is direct opportunity to redefine existing containment research within the scope defined through the single domain ontology outlined in this study. 

\section*{References}
\sloppy
Agar, N. (2016). Don’t Worry about Superintelligence. Journal of Evolution and Technology. 26(1): 73-82.  

Amodei, D., Olah, C., Steinhardt, J., Christiano, P., Schulman, J., \& Mane, D. (2016). Concrete Problems in AI Safety. arXiv. Retrieved from arXiv:1606.06565
 
Babcock, J., Kramar, J., \& Yampolskiy, R. (2016). The AGI Containment Problem. Lecture Notes in Artificial Intelligence 9782 (AGI 2016, Proceedings) 53-63.
 
Babcock, J., Kramar, J., \& Yampolskiy, R. (2017). Guidelines for Artificial Intelligence Containment. ArXiv Preprint. doi:1707.08476
 
Baillie, J-C. (2016). Why AlphaGo Is Not AI. IEEE Spectrum. Retrieved 
from https://spectrum.ieee.org/automaton/robotics/artificial-intelligence/why-alphago-is-not-ai
 
Baum, S., Goertzel, B., \& Goertzel, T. (2011). How long until human-level AI? Results from an expert assessment. Technological Forecasting and Social Change, 78(1). 185-195. 
 
Bishop, M. (2009). Why Computers Can’t Feel Pain. University of London. Retrieved from http://citeseerx.ist.psu.edu/viewdoc/download?doi:10.1.1.492.3886\&rep=rep1\&type=pdf  
 
Blanco, C., Lasheras, J., Valencia-Garc, R., Fern, E., Toval, A., \& Piattini, M. (2008). A Systematic Review and Comparison of Security Ontologies. IEEE: 813-820. doi:https://doi.org/10.1109/ARES.2008.33
 
Bostrom, N. (2002). Existential Risks: Analyzing Human Extinction Scenarios and Related Hazard. Journal of Evolution and Technology, 9(1). 
 
Bostrom, N. (2014). Superintelligence: Paths, dangers, strategies. Oxford: Oxford University Press.  
 
Bostrom, N., \& Cirkovic, M. (2011). Global Catastrophic Risk. Oxford: Oxford University Press.  
 
Bostrom, N., \& Yudkowsky, E. (2011) The Ethics of Artificial Intelligence. Cambridge Handbook of Artificial Intelligence, eds. Ramsey W., \& Frankish, K. Cambridge: Cambridge University Press.
 
Chalmers, Awret, \& Appleyard. (2016). The Singularity: Could artificial intelligence really out-think us (and would we want it to)? Journal of Consciousness Studies: Imprint Academic.  
 
Denil, M., Agrawal, P., Kulkarni, T., Erez, T., Battaglia, P., \& de Freitas, N. (2017). Learning to perform physics experiments via deep reinforcement learning.  Under review as a conference paper to ICLR. Retrieved from https://arxiv.org/pdf/1611.01843v1.pdf 
 
Grau, C. (2006). There Is No ‘I’ in ‘Robot’: Robots and Utilitarianism. IEEE Intelligent Systems 21(4): 52–55. doi:10.1109/MIS.2006.81
 
Guo, W., \& Aarabi. (2016). Hair Segmentation Using Heuristically-Trained Neural Networks. IEEE Transactions on Neural Networks and Learning Systems, 1-12. doi:10.1109/tnnls.2016.2614653
 
Horowitz, M. (2014). “The Looming Robotics Gap,” Foreign Policy. Retrieved from 
http://www.foreignpolicy.com/articles/2014/05/05/the\_looming\_robotics\_gap\_us\_militay\_technology\_dominance
 
Kott, A. (2015). Science of Cyber Security as a System of Models and Problems. New York: Springer. 
 
Kurzweil, R. (2005). The Singularity Is Near: When Humans Transcend Biology. New York: Penguin.
 
Le, Q., \& Schuster, M. (2016). A Neural Network for Machine Translation, at Production Scale. Google. Retrieved from https://research.googleblog.com/2016/09/a-neural-network-for-machine.html
 
Lewis \& Modirzadeh. (2016). War-Algorithm Accountability. Harvard Law School Program on International Law and Armed Conflict, Harvard Law Review. Retrieved from https://papers.ssrn.com/sol3/papers.cfm?abstract\_id\=2832734 
 
Longstaff, T. (2010). Cyber science: moving from the toes to the shoulders of giants. In 
Proceedings of the Sixth Annual Workshop on Cyber Security and Information Intelligence Research, 10. Retrieved from http://dl.acm.org/citation.cfm?id=1852677
 
Ma, B., Nahal, S., \& Tran, F. (2015). Thematic Investing: Robot Revolution Global Robot \& AI Primer. Bank of America/Merrill Lynch. Retrieved from https://olui2.fs.ml.com/publish/content/application/pdf/GWMOL/RIC-Report-January-2016.pdf
 
Martin, P. Y., \& Turner, B. A. (1986). Grounded theory and organizational research. The journal of applied behavioral science, 22(2), 141-157.
 
Maxion, R. A., Longstaff, T. A., \& McHugh, J. (2010). Why is there no science in cyber science?: a panel discussion at NSPW 2010. In Proceedings of the 2010 workshop on New security paradigms, 1–6. ACM. Retrieved from http://dl.acm.org/citation.cfm?id=1900548
 
McDaniel, P., Rivera, B., \& Swami, A. (2014). Toward a science of secure environments. IEEE Security \& Privacy, 12(4), 68-70.
 
Müller, V. (2014). Risks of artificial general intelligence. Journal of Experimental and Theoretical Artificial Intelligence, 26(3). 
 
Nilsson, N. (2009). The Quest for Artificial Intelligence: A History of Ideas and Achievements, Cambridge: Cambridge University Press. 
 
National Science and Technology Council. (2016). PREPARING FOR THE FUTURE OF ARTIFICIAL INTELLIGENCE OF ARTIFICIAL INTELLIGENCE. Executive Office of the President.   
 
Ning, H., Li, Q., Wei, D., Liu, H., \& Zhu, T. (2017). Cyberlogic Paves the Way From Cyber Philosophy to Cyber Science. IEEE Internet of Things Journal, 4(3), 783–790. doi: https://doi.org/10.1109/JIOT.2017.2666798
 
Oltramari, A., Cranor, L. F., Walls, R. J., \& McDaniel, P. D. (2014). Building an Ontology of Cyber Security. In STIDS, 54–61.
 
Poli, R. (1996). Ontology for knowledge organization. Advances in Knowledge Organization, 5, 313–319.
 
Powers, T. (2006). Prospects for a Kantian Machine. Intelligent Systems, IEEE, 21, 46-51. doi:10.1109/MIS.2006.77 
 
Ren, F. (2012). A demo for constructing domain ontology from academic papers. Proceedings of COLING 2012: Demonstration Papers, 369–376.
 
Silver, D., Huang, A., Maddison, C., Guez, A., Sifre, L., van den Driessche, G.,… Hassabis, D. (2016). Mastering the game of Go with deep neural networks and tree search. Nature, 529, 484-489.

Sotala, K. \& Yampolskiy, R. (2016). Responses to catastrophic AGI risk: a survey. Physics Scripta 90, 1-33. doi:10.1088/0031-8949/90/6/069501 
 
Stone, P., Brooks, R., Brynjolfsson, R., Calo, R., Etzioni, O., Hager, G.,… Teller, A. (2016). 
“Artificial Intelligence and Life in 2030: One Hundred Year Study on Artificial 
Intelligence: Report of the 2015-2016 Study Panel”, Stanford University. Retrieved from http://ai100.stanford.edu/2016-report 
 
Subhashini, R. \& Akilandeswari, J. (2011). A survey on ontology construction methodologies. International Journal of Enterprise Computing and Business Systems, 1(1), 60–72.
 
Suddaby, R. (2006). From the editors: What grounded theory is not. Academy of management journal, 49(4), 633-642.
 
Swartout, B., Patil, R., Knight, K., \& Russ, T. (1997). Toward Distributed Use of Large-Scale Ontologies. Symposium on Ontological Engineering, Association for the Advancement of Artificial Intelligence - AAAI.
 
Uschold, M., \& King, M. (1995). Towards a methodology for building ontologies. Artificial Intelligence Applications Institute, University of Edinburgh. Retrieved from http://www.aiai.ed.ac.uk/project/oplan/documents/1995/95-ont-ijcai95-ont-method.pdf
 
Wissner-Gross, A., \& Freer, C. (2013). Causal Entropic Forces. Physical Review Letters 110(168702). 
 
Yampolskiy, R. V. (2014). Utility Function Security in Artificially Intelligent Agents. Journal of Experimental and Theoretical Artificial Intelligence (JETAI): 1-17.
 
Yampolskiy, R. V. (2016). Taxonomy of Pathways to Dangerous Artificial Intelligence. AAAI Conference on Artificial Intelligence AI, Ethics, and Society: Technical Report WS- 16-02.  
 
Yampolskiy, R. V., \& Fox, J. (2012). Safety Engineering for Artificial General Intelligence. Springer Institute for Artificial Intelligence 32, 217-226. doi:10.1007/s11245-012-9128-9. 

Yudkowsky, E. (2008). Artificial Intelligence as a Positive and Negative Factor in Global Risk. In Global Catastrophic Risks, ed. Bostrom, N. and Cirkovic, M., New York: Oxford University Press.

\end{document}